\documentclass{article}
\usepackage{subcaption}
\usepackage{booktabs}
\usepackage{graphicx} 
\usepackage{qtree}
\usepackage{authblk}
\usepackage{listings}

\title{Hierarchical Delay Attribution Classification using Unstructured Text in Train Management Systems\footnote{This research has been performed with funding from Swedish Transport Administration (Trafikverket) via the research program Capacity in the Railway System (www.kajt.org). Corresponding author: anton.borg (at) bth.se}}
\author[1]{Anton Borg}

\author[2]{Per Lingvall}

\author[3]{Martin Svensson}

\affil[1]{Department of Computer Science, Blekinge Institute of Technology, Sweden}

\affil[2]{Swedish Transport Administration, Sweden}

\affil[3]{Department of Industrial Economics, Blekinge Institute of Technology, Sweden}
\date{}                     
\begin{document}

\maketitle

\begin{abstract}
EU directives stipulate a systematic follow-up of train delays. In Sweden, the Swedish Transport Administration registers and assigns an appropriate delay attribution code. However, this delay attribution code is assigned manually, which is a complex task. In this paper, a machine learning-based decision support for assigning delay attribution codes based on event descriptions is investigated. The text is transformed using TF-IDF, and two models, Random Forest and Support Vector Machine, are evaluated against a random uniform classifier and the classification performance of the Swedish Transport Administration. Further, the problem is modeled as both a hierarchical and flat approach. The results indicate that a hierarchical approach performs better than a flat approach. Both approaches perform better than the random uniform classifier but perform worse than the manual classification.
\end{abstract}

\section{Introduction}
EU-directives stipulate a systematic follow-up of train delays ~\cite{Trafikverket2020, Trafikverket2023}. In Sweden, the follow-up is managed by the Swedish Transport Administration~\cite{Trafikverket2023}, who registers and assigns an appropriate delay attribution code to the related train delay~\cite{JOBORN2022100339}. The delay registrations and their corresponding delay attribution code are used to quantify the quality of the Swedish railway system over time, but also to identify accountability of the actor that has caused the disturbance and hence the delay. In turn, the part who is decided to be accountable for the delay may have to pay a fee, termed "quality fee". The challenges with the delay registration is to determine the appropriate delay attribution code, which is a subject of recurring follow-ups where lessons learned, statistics and conclusions are part of a continuous quality improvement process (see for example ~\cite{Trafikverket2023}and ~\cite{Trafikverket2020}

The process to determine the delay attribution code involves multiple actors and is distributed over several days and across multiple information systems. Determining the delay attribution code is difficult for three reasons. First, there currently exists approximately $200$ potential delay attribution codes in a three level hierarchical structure. Second, the data might be incomplete when the initial classification is conducted. Third, the data needs to be manually collected and documented, containing domain specific syntax. 

When notified about a delay, the dispatcher needs to make the necessary investigations to identify the cause for the delay, e.g. by contacting the train driver, contacting entrepreneurs doing maintenance work, or by looking for additional information in relevant and available systems. In addition to assigning the delay a Disturbance code, the train dispatcher also needs to provide a short description of the situation that can be used later in the process.

However, the dispatcher have limited time available to carry out this investigation, in addition to existing tasks, thus contributing to an already intensive workload for traffic control staff. Further, given that train dispatchers are geographically separated from actual event it might be difficult to acquire timely, accurate, or complete information. 

The initial delay attribution coding is therefore reviewed, during the following three days, by delay attribution code experts. When these experts have reviewed, revised or confirmed the existing code, it is released to the companies operating the trains. If the train company has any objections or additional information about the case they need to report back before the end of 8th day following the event. The final delay attribution code, and hence it's corresponding quality fee, is then set in the system by the end of the 9th day following the event.

Consequently, the registration of train delays are today a manual process requiring dispatchers to gather the initial information necessary and manually setting the delay attribution code for the delay down to the at least the 2nd level in the hierarchy~\cite{JOBORN2022100339}. This can be time consuming and error-prone~\cite{mullenbach-etal-2018-explainable}, as well as dependent on the experience of the dispatcher. Further, there is a large decision space ~\cite{simon1990invariants,simon2000theories} for dispatchers to navigate, which can lack information, and contain context-sensitive and ambiguous information, making automation more difficult.

However, classification of train delays could be assisted by automated solutions~\cite{mullenbach-etal-2018-explainable}. Given the textual information provided by dispatchers it should be possible to provide a decision support system to assist dispatchers in the delay registration process, thus reducing their workload. A machine learning based decision support system for setting the delay attribution code would also provide a secondary input for the dispatchers to use, thus potentially reducing classification errors. 

Given that the decision space spans multiple levels and delay attribution codes within each level, we model this problem as a hierarchical multi-class text-classification problem. Hierarchical solutions have been suggested to perform better than flat approaches in other domains~\cite{sebastiani2002machine, perotte2014diagnosis}. As such, the problem is investigated with a systematic classification process across multiple levels, where models are trained using data sampled based on the parent code presence. This hierarchical approach enables the classification of instances into increasingly specific categories, allowing for a more nuanced and accurate classification process. 

As such, given the unstructured text for each delay event, this paper aims at investigating the possibility to automate the classification of delay attribution codes to at least the second level, in order to provide decision support to the involved actors in the Swedish railway industry. The investigation is limited to data available in the unstructured text of the reports, and as such does not involve additional information that operators have available but have chosen to not include in the reports. 

The paper is outlined as follows. Related work is presented in Section~\ref{sec:relwork}, followed by the methodology in Section~\ref{sec:method}. The methodology contains a description of the data and preprocessing, the algorithms and evaluation metrics used, as well as the experiment setup and statistical evaluation. Section~\ref{sec:res} presents the results of the experiments and the statistical analysis. Finally, the results are discussed in Section~\ref{sec:disc}, and the conclusions and future work presented in Section~\ref{sec:con}.

\section{Related Work}
\label{sec:relwork}
While there has been some research into delay predictions of trains~\cite{JOBORN2022100339, SPANNINGER2022100312, TIONG2023104027, 9140377, laifa2021train, arshad2019prediction, liu_prediction_2023}, as well as  other forms of travels~\cite{yi2021flight, mamdouh2024improving, sirisati2021aviation, alshaer2019efficient}, not much research has been concerned with classifying the different types of delays. Various factors can give rise to delays, resulting in disruptions of varying duration, and which have vastly different mitigation solutions~\cite{konig_review_2020}. Further, the information describing the delay may be incomplete, making instant classification of the delay type difficult~\cite{konig_review_2020}. 

Machine Learning-based text classification has been successfully used in several domains to categories text documents according to their content~\cite{sebastiani2002machine, kowsari2019}. Early contributions laid foundational principles for employing traditional techniques like Term Frequency-Inverse Document Frequency (TF-IDF) and Support Vector Machines (SVM) in text classification tasks~\cite{joachim1998, russell2010artificial}. Recently, Deep Learning techniques has significantly impacted this domain showcasing the efficacy of Deep Learning approaches for text classification~\cite{kim2014, borg_e-mail_2021}. 
Automated text classification has been shown to work in specialized domains, such as medical text classification~\cite{mullenbach-etal-2018-explainable}, and email classification~\cite{borg_e-mail_2021}. Such domains can have unstructured text that are different from other documents, e.g. emails have been argued to be a separate type of documents~\cite{BARON1998133}. 

For tasks which involves a large number of potential classes, hierarchical approaches~\cite{sebastiani2002machine, kowsari2019, perotte2014diagnosis} improves the accuracy and efficiency of text classification models. Hierarchical approaches streamline the classification process by breaking it down into manageable steps, making it more efficient, especially when dealing with a large number of potential categories. 

While research into delay predictions for railway traffic and airlines seem to be an emerging research area, there does not seem to be any research into classifying the different types or reasons for the delay. The closest study focused on using text data for delay prediction~\cite{liu_prediction_2023}. Classifying different types of delays and, by extension, the reasons for the delay is an important step towards mitigating delays. By utilizing the delay reports that are written by operators together with the existing hierarchical delay attribution codes, it should be possible to automate the delay classification. This is supported by the findings from ICD classifications~\cite{perotte2014diagnosis}. Similar to emails~\cite{borg_e-mail_2021} and medical documents~\cite{mullenbach-etal-2018-explainable}, the unstructured text of the delay reports can be considered a distinct type of documentations as they, at least in this case, are written using domain specific terminology and syntax. Further, the texts in this study are also written in Swedish.

\section{Method}
\label{sec:method}
This section describes the data collection and preprocessing, the algorithms, evaluation metrics, experiment setup, and statistical tests used.
\subsection{Data and Data Preprocessing}
\label{sec:data}
The data used in this study were provided by Swedish Transport Administration and contains internal unstructured text about the cause of the delay as well as the code for the delay. The features of the data can be seen in Table~\ref{tab:features}.
As previously stated there are $\approx200$ delay attribution codes grouped into a three level hierarchy, with five top level codes, multiple refinement codes on the second level for the first level code, and additional refinement codes on the third level for the level 2 code~\cite{JOBORN2022100339}.   

  \begin{description}
      \item[Operational Management (D)]: disruptions caused by, for example, the Swedish Transport Administration's own prioritization, incorrect handling or incorrect traffic information.
      \item[Consequential cause (F)]: disturbances caused by, for example, an expected connection, round-trip times or lack of available track.
      \item[Infrastructure (I)]: damage to signaling and electrical installations, track and track switches as well as culverts, tunnels and bridges. This also includes disturbances caused by track works and weather phenomena such as solar curves.
      \item[Railway company (J)]: disruptions caused by, for example, locomotive and machine faults, terminal and platform management or vehicle or staff shortages.
      \item[Accidents/incidents and external factors (O)]: disruptions caused by, for example, weather, unauthorized persons on tracks, accidents, sabotage and trains that arrive late to Sweden from other countries.
  \end{description}
      
Thus, the reason for a delay ranges from the trains' breaking system, lack of staff operating the trains, to environmental circumstances, such as storms, snow or rain. After deciding on the level one delay attribution code (D,F,I,J or O), additional letters may be added for a level two code (such as JTP or FAT), and numbers for a level three code, further specifying the code. The total number of codes the dispatcher may choose from exceeds 200 codes.

The dataset was collected during 30 days in Maj-June 2023, containing $34901$ instances. The data was collected from the whole of Sweden and consists of events impacting trains during these days. This dataset shares similarities with a prior dataset~\cite{JOBORN2022100339}, with a key distinction being the inclusion of data points for each day over the 10-day process. This inclusion enhances the dataset's capacity to identify variations and changes in the collected data over time, particularly concerning the delay attribution code and unstructured text. This study primarily focuses on day 0 and day 10. It should also be noted that not all possible codes are available in the collected data. 

Each row contains an event, with the delay attribution code, and unstructured free-text explaining what happened and consequences. The delay attribution code is described as both condensed and verbose form (feature $n1, n2, n3$). In the latter feature $n1$ denotes level one of the delay attribution code and corresponds to the first letter of the condensed form, feature $n2$ denotes level two of the delay attribution code and corresponds to the next two letters of the condensed form, and feature $n3$ denotes level three of the delay attribution code and corresponds to the last characters of the condensed form. The delay attribution code registered on the same day as the event and on day 10 after the event is available in the dataset, in accordance with the process used.

The dataset was preprocessed by removing duplicate entries, or entries with no free-text, and only keeping rows where the label occurred at least 100 times. This was done in order to have adequate data to train and evaluate for each class~\cite{flach2012machine}. This resulted in a dataset size of $21484$ instances.

Further, punctuation and line breaks were removed from the free-text, and the text transformed to lowercase. The term \emph{sth} is used throughout the texts, but written differently; \lstinline{sth <speed>}, \lstinline{sth<speed>}, \lstinline{sth <speed>km}, or similar variations. To address this inconsistency, we remove the spacing variations, ensuring a standardized format, \lstinline{sth<speed>}, where the numbers are connected to the \emph{sth} keyword.
Individual train identifiers occurs frequently in the free-text, and in some cases only train identifiers are entered. These were replaced with a place holder, \emph{TRAINNR}. There are 8032 instances where the text consists only of train identifiers. This indicates that there is substantial knowledge not documented in the unstructured text.

\begin{table}[ht]
    \footnotesize
    \begin{tabular}{ll}
         \toprule
            Feature & Description \\
         \midrule
         $eventcode$ & Event identification, e.g. 439394\\
         $text$ & Free-text describing the event and consequences\\
         $label$ & delay attribution code for day 0, e.g. DPR 03\\
         $n1_0$ & Day 0 first level code, e.g. D\\ 
         $n2_0$ & Day 0 second level code, e.g. PR\\
         $n3_0$ & Day 0 third level code, e.g. 03\\
         $n1_{10}$ & Day 10 first level code, e.g. J\\
         $n2_{10}$ & Day 10 second level code, e.g. PR\\
         $n3_{10}$ & Day 10 third level code, e.g. 05\\
         \bottomrule
    \end{tabular}
    \caption{Description of data in Dataset 1 and 2. Please note that the label for day 0 might differ from day 10.}
    \label{tab:features}
\end{table}

The free-text was, during the experiment, tokenized using TF-IDF. TF-IDF is a useful, but simple way of transforming free-text into a structured format useful for machine learning algorithms~\cite{russell2010artificial}. It has shown to enable text classification with a high performance~\cite{sebastiani2002machine}. While newer natural language processing methods involve using transformers or deep-learning to capture context and understanding better, TF-IDF is an reasonable approach for initial evaluations~\cite{russell2010artificial}. In the experiment, stop-words were removed, the texts were transformed into 1-, 2-, and 3-grams based on the words of the texts, and the top 1000 features ordered by term frequency across the corpus were kept.

\subsection{Algorithms}
\label{sec:algo}
Random Forest (RF) is an ensemble learning method that operates by constructing multiple decision trees during training and outputs the class that is the majority decision of the individual trees~\cite{flach2012machine}. Each tree in the ensemble is built from a random sample of the training data using a randomly selected subset of features, and the final prediction is determined by aggregating the individual tree predictions. This approach helps mitigate overfitting and improves the model's generalizability, making Random Forest a versatile and widely used algorithm for various machine learning tasks, including classification~\cite{cutler}. The ability of Random Forest to handle high-dimensional data,  and effectively handle outliers and missing values contributes popularity in different domains~\cite{kowsari2019}.

Support Vector Machines (SVM) has shown to be a popular and well performing algorithm when it comes to text-classifications~\cite{flach2012machine}. Support Vector Machines (SVM) is a powerful supervised learning algorithm used for classification and regression tasks. In the context of classification, SVM aims to find the optimal hyperplane that separates different classes by maximizing the margin, i.e., the distance between the hyperplane and the nearest data points of each class. SVM is capable of handling complex data distributions, making it a versatile and effective tool for various machine learning tasks, including text categorization~\cite{flach2012machine,kowsari2019, sebastiani2002machine}. 

Given that some of the classes are rarely occurring, conformal prediction is used to improve the classification results by estimating the uncertainty and removing predictions that are uncertain~\cite{angelopoulos2021gentle}. Conformal Prediction is a framework in machine learning that provides a method for assigning reliable measures of confidence to individual predictions~\cite{vovk2005algorithmic}. By constructing prediction regions that accommodate a predefined significance level, Conformal Prediction enables the assessment of the credibility of predictions, allowing for a quantification of the uncertainty associated with the predictions made by machine learning models. This framework is particularly useful in enhancing the reliability of predictions and has found applications in various domains, including text-classification, health-care, and computer vision, where the accurate assessment of prediction credibility is of utmost importance in decision-making processes~\cite{angelopoulos2021gentle}.

Uniform Classifier is a simple baseline model often used for comparison in machine learning tasks. It randomly predicts class labels according to the distribution of the training data, without learning any pattern from the input features. This approach is particularly useful for assessing the performance of other more complex models, serving as a reference point for evaluating the effectiveness of the applied machine learning techniques. 

In this study the algorithms are implemented using scikit-learn library for Python~\cite{scikit-learn}.

TKL is the optimal performance and is based on the manual classification effort. This is done by computing the F1-score using the true labels for day 0 and day 10, giving an indication on how well the operatives are capable of classifying the delay attribution codes on day 0. The F1-score is described in more detail in the following section.

\subsection{Experiment setup}
\label{sec:exp}
Our experiment evaluates two approaches. The first is whether we can classify the text with regard to the third level delay attribution code. In this scenario, the text is transformed using TF-IDF, the algorithm trained with the processed text and the third level delay attribution code as class labels. A 10-fold cross conformal validation strategy is used, i.e. an ordinary 10-fold cross validation, but the training set is split equally into a training and calibration set. The model is trained on the remaining training data, calibrated using the calibration set, and finally the model is evaluated on the test set. The experiment is evaluated using the F1-score. The F1-Score is the harmonic mean of precision and recall, where precision is TP (TP) / (TP + FP) and recall is TP / (TP + FN) given that TP is true positives, FP is false positives, and FN is false negatives. As such, the F1-score is calculated as per the following equation:
\begin{equation} 
    F1 = 2 * \frac{precision * recall}{precision + recall}
\end{equation}
 In a multi class setting this is calculated per class and averaged.

In the second approach, delay attribution codes are classified at varying levels of granularity using the verbose delay attribution codes represented as $n1$, $n2$, and $n3$. A hierarchical classification system comprising three levels is implemented, enabling classification of instances into increasingly specific categories, allowing for a more nuanced and accurate classification process~\cite{sebastiani2002machine, kowsari2019}. This is similar to Stacking SVM~\cite{kowsari2019, sun2001hierarchical}. However, instead of having binary-class classifiers for each leaf, multi-class classifiers are used for each node. This approach is utilized for both the SVM and the Random Forest algorithm.
\begin{figure}[ht]
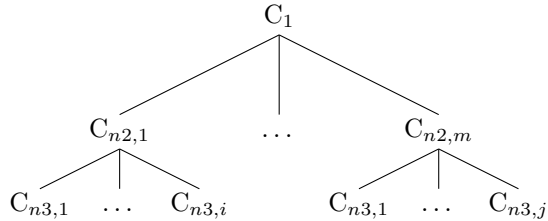

    \Tree [.C$_{1}$ [ C$_{n3,1}$ {\ldots} C$_{n3,i}$ ].C$_{n2,1}$ {\ldots} [ C$_{n3,1}$ {\ldots} C$_{n3,j}$ ].C$_{n2,m}$ ]   
    \caption{The Hierarchical classification approach visualized. Each node C is a multi-class classifier, trained using a specific delay attribution code and its sub-classes as classification targets.}
    \label{fig:hierarch_approach}
\end{figure}

At the initial level, classification is attempted based on $n1$ class labels. This involves transforming the text through TF-IDF, training an algorithm with the processed text, and utilizing $n1$ codes recorded on day 0 (referred to as feature $n1_0$) as class labels. Model evaluation is carried out using 10-fold cross-validation, with an equal division of the training set into training and calibration subsets. The model is trained on the remaining training data, and predictions are calibrated using the calibration set (i.e. a  cross-conformal approach). Performance assessment employs the previously outlined evaluation metric and includes evaluation against both the $n1$ codes recorded on day 0 and day 10 (i.e., features $n1_0$ and $n1_{10}$).

In the second level of the classification hierarchy, a comparable approach to the first level is applied. However, in this context, a distinct model is constructed for each class within $n1_0$. Data samples are extracted for each class within $n1_0$, and the corresponding labels from $n2_0$ are used as target labels. Given that $n1_0$ contains four distinct target labels, this approach results in the creation of four distinct models, each tailored to one of the four data-sets.

Advancing to the third level, a similar evaluation methodology is retained. For every class within $n2_0$, data samples are collected, and a model is developed using $n3_0$ as target labels. Similarly to before, performance evaluation entails assessing this model against both $n3_0$ and $n3_{10}$ target labels. This multi-level approach enables a progressive classification of delay attribution codes, accommodating increasing levels of specificity.

\subsection{Statistical evaluation}
\label{sec:stat}
When evaluating the difference between a flat versus a hierarchical approach, Kruskal-Wallis and the Conover post-hoc test is used to determine where, if any, statistical significance is manifested ~\cite{sheskin2003handbook}. 

For evaluating the performance of the approaches for the classification of delay attribution codes on Level 1, Kruskal-Wallis is used to investigate whether there is a statistical significant difference between the algorithms for the day 0 and day 10. Conover post-hoc test is used to determine how the difference is manifested~\cite{sheskin2003handbook}. 

For the classification of delay attribution codes on Level 2 and 3, the different delay attribution codes investigated are seen as different data-sets (e.g. D, I, J, O for level 2). As such, Friedman's test is used to investigate whether there is a statistical significant difference between the algorithms, and a nemenyi post-hoc test to determine how the difference is manifested between the approaches~\cite{demvsar2006statistical, sheskin2003handbook}.

Friedman's Test ranks based on the mean evaluation metrics and is, consequently more conservative compared to Kruskal-Wallis. It is expected that Kruskal-Wallis will detect more differences than Friedman's test is capable of~\cite{sheskin2003handbook}.

\section{Results}
\label{sec:res}
The ability to correctly classify the delay attribution code on different levels is investigated in this study. In the following subsections the results of the experiments are presented. The results of the statistical analysis are also presented in this section.

\subsection{Flat vs Hierarchical classification}
\label{sub:flatvHier}
Comparing a flat vs hierarchical classification approach indicates that the hierarchical approach performs better on level two and three. This can be observed in Table~\ref{tab:f_vs_h}. The algorithms where investigated in a hierarchical and flat approach. In this scenario, classification on level one between the two approaches can be considered equivalent and as such is not interesting to compare.

While the hierarchical approach in general have a higher F1-score than compare to flat approach, the standard deviation is also higher. This indicates that there are certain classes that have a higher performance than the mean, but also lower. This is not unexpected as the results for the hierarchical approach is the mean over the different classes, which can be seen in \ref{sec:level3} to have classes that are more difficult to classify. 

For the Random Forest and SVM algorithms, the hierarchical approach performs better than the flat approach. However, the hierarchical approach have a higher standard deviation indicating that certain classes might be more difficult to classify than others. For Random Forest at Level 2, the hierarchical approach outperforms the flat approach (F1-score of $0.860 \pm 0.102$ vs. $0.777 \pm 0.010$), and at Level 3, the hierarchical approach maintains a competitive performance (F1-score of $0.733 \pm 0.113$ vs. $0.708 \pm 0.009$). SVM performs similarly at Level 2, where the hierarchical approach outperforms the flat approach (F1-score of $0.876 \pm 0.100$ vs. $0.797 \pm 0.008$). While at Level 3, a similar trend is observed, with the hierarchical approach showing improved performance (F1-score of $0.737 \pm 0.114$ vs. $0.723 \pm 0.009$).

Looking at the uniform classifier, it is clear that the solution space for the hierarchical models are much smaller than for the flat approach. Since the uniform classifier randomly sets the class the score should be approximately $1/|c|$ where $c$ is the number of classes in the solution space. At Level 2, the hierarchical approach significantly outperforms the flat approach (F1-score of 0.335 vs. 0.075). And at Level 3, a similar pattern is observed, with the hierarchical approach showing improved performance (F1-score of 0.313 vs. 0.023). However, the standard deviation for the hierarchical solution is much higher, indicating that there is a greater variance in the size of the solution space for the hierarchical solution, which is expected.

\begin{table*}[ht]
    \centering
    \footnotesize
    \begin{tabular*}{\textwidth}{@{\extracolsep\fill}llrr@{}}
        \toprule
        Algorithm &  &   level 2     & level 3\\
        \midrule
        Uniform Classifier   & Flat      & $0.075$ ($0.004$) & $0.023$ ($0.002$)  \\
                & Hierarchical   & $0.335$ ($0.113$) & $0.313$ ($0.157$)  \\
        Random Forest      & Flat      & $0.777$ ($0.010$) & $0.708$ ($0.009$)  \\
                & Hierarchical   & $0.860$ ($0.102$) & $0.733$ ($0.113$)  \\
        SVM     & Flat      & $0.797$ ($0.008$) & $0.723$ ($0.009$)  \\
                & Hierarchical   & $0.876$ ($0.100$) & $0.737$ ($0.114$)  \\
        TKL     & Flat      & $0.972$ ($0.004$) & $0.962$ ($0.004$)  \\
                & Hierarchical   & $0.963$ ($0.034$) & $0.922$ ($0.066$)  \\
        \bottomrule
    \end{tabular*}
    \caption{Mean F1-score (and standard deviation) per algorithm for the two different approaches.}
    \label{tab:f_vs_h}
\end{table*}

\begin{figure}[ht]
    \centering
    \subcaptionbox{Second Level\label{fig:sp_hvf_n2}}
       {\includegraphics[width=0.4\textwidth,keepaspectratio]{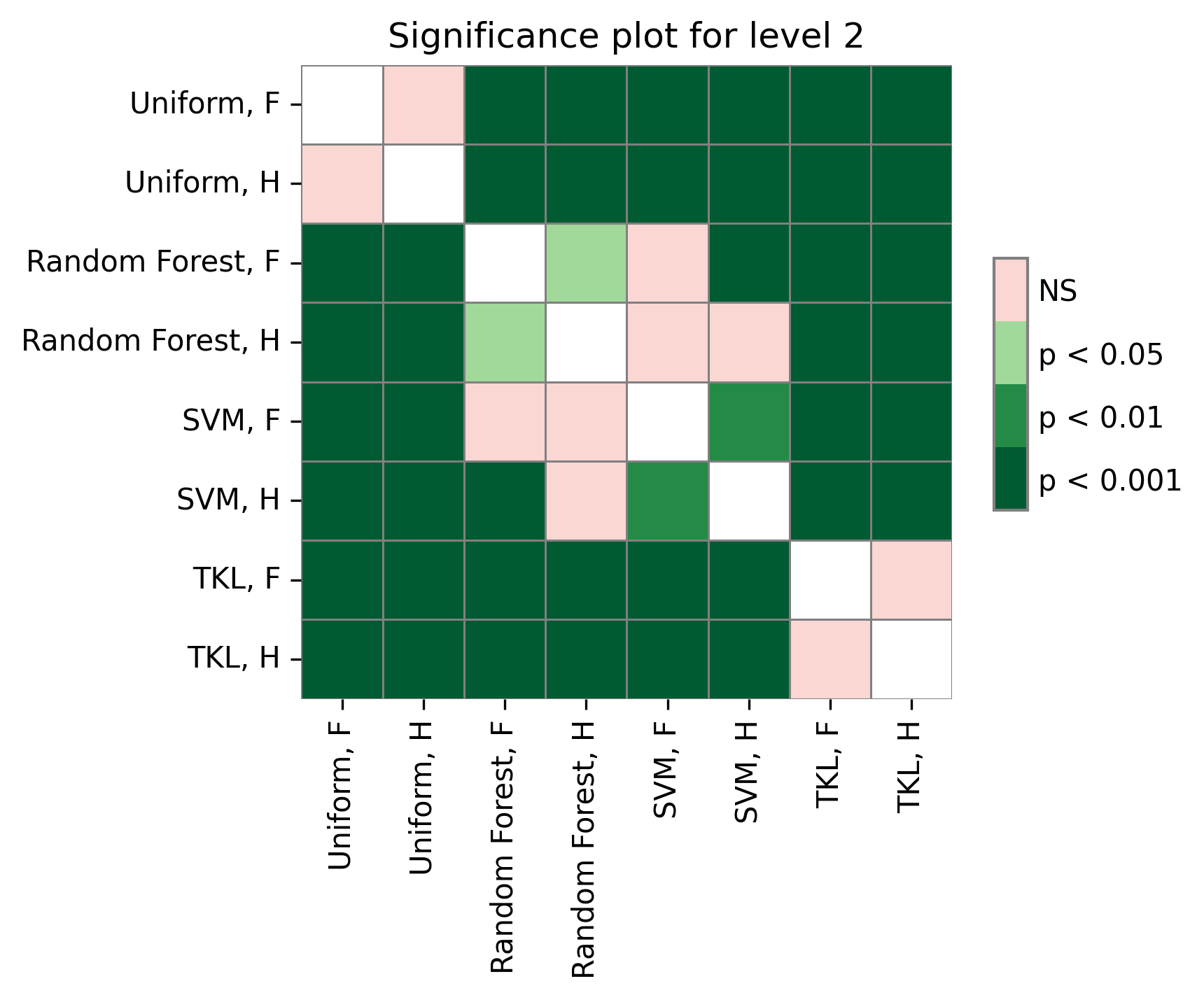}}
    \subcaptionbox{Third level\label{fig:sp_hvf_n3}}
        {\includegraphics[width=0.4\textwidth,keepaspectratio]{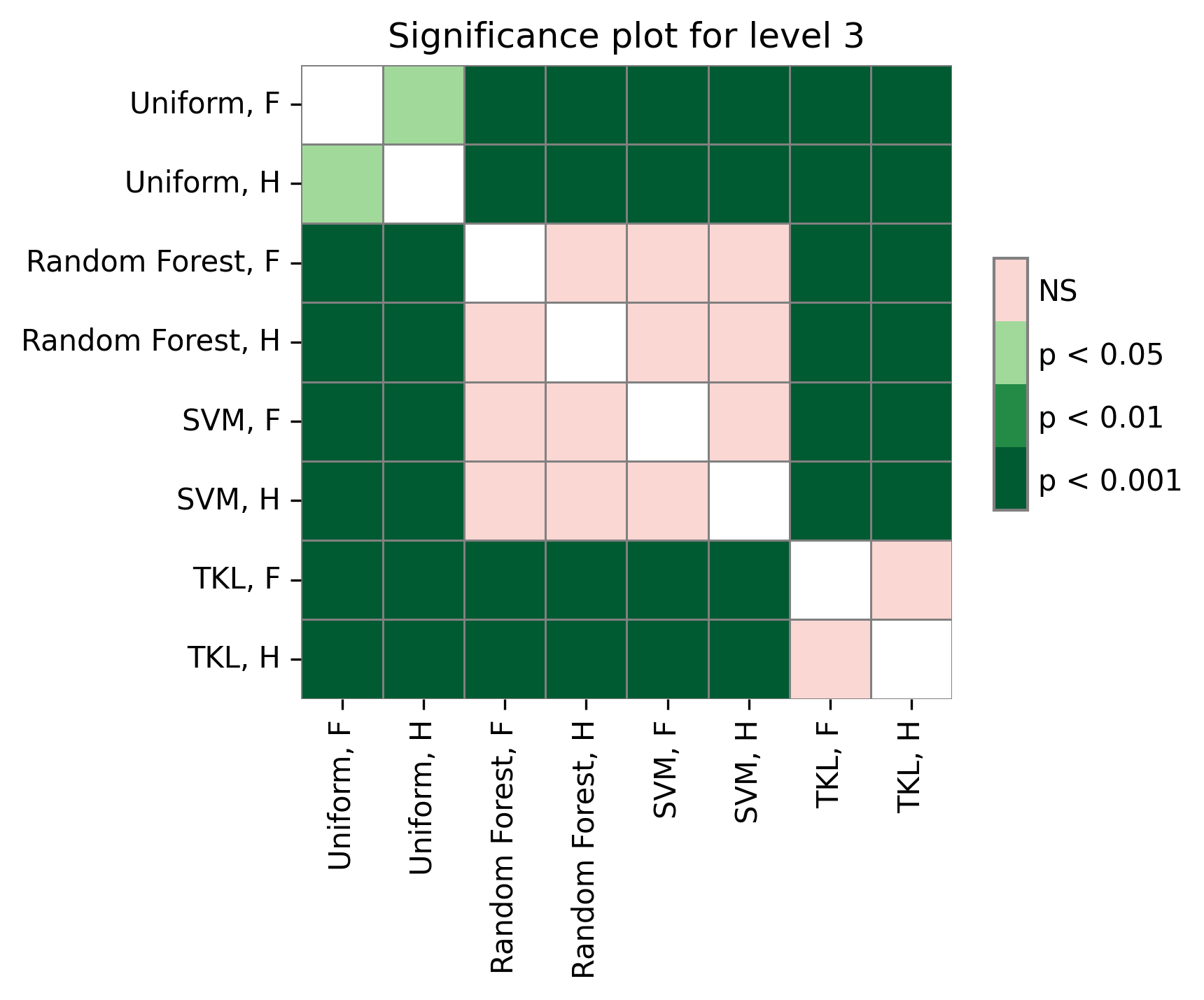}}
    \caption{Significance plot comparing the results for a flat (, F) approach against an Hierarchical (, H) approach.}
    \label{fig:sp_aproaches}
\end{figure}

A Kruskal-Wallis test showed that at there was a significant difference of means for both the third level ($H = 337.52$, $p<0.001$) and the second level ($H = 154.1359$, $p<0.001$). A Conover's post-hoc test indicates that the difference is a statistical significance between the uniform classifier and the other approaches($p<0.001$), see Figure~\ref{fig:sp_aproaches}. There is a statistical significant difference between Random Forest flat and Hierarchical ($p<0.05$) and the hierarchical SVM ($p<0.001$). The test doesn't find any statistical significant difference between the flat SVM and the Random Forest approaches. There does also seem to be a significant difference between the flat and the hierarchical approach for the SVM algorithm ($p<0.01$). This indicates that the hierarchical approach performs statistically significant better than the flat approach.

A sidenote when comparing the two approaches is that the hierarchical approach finishes in half the time that the flat approach takes, $\approx 8$ minutes vs $\approx 16$ minutes. At least in this experiment setting.

\subsection{Level 1}
\label{sec:level1}

\begin{figure}[ht]
    \centering
    \subcaptionbox{Day 0 and Day 10\label{fig:f1_n1_d0}}
       {\includegraphics[width=0.2\textwidth,keepaspectratio]{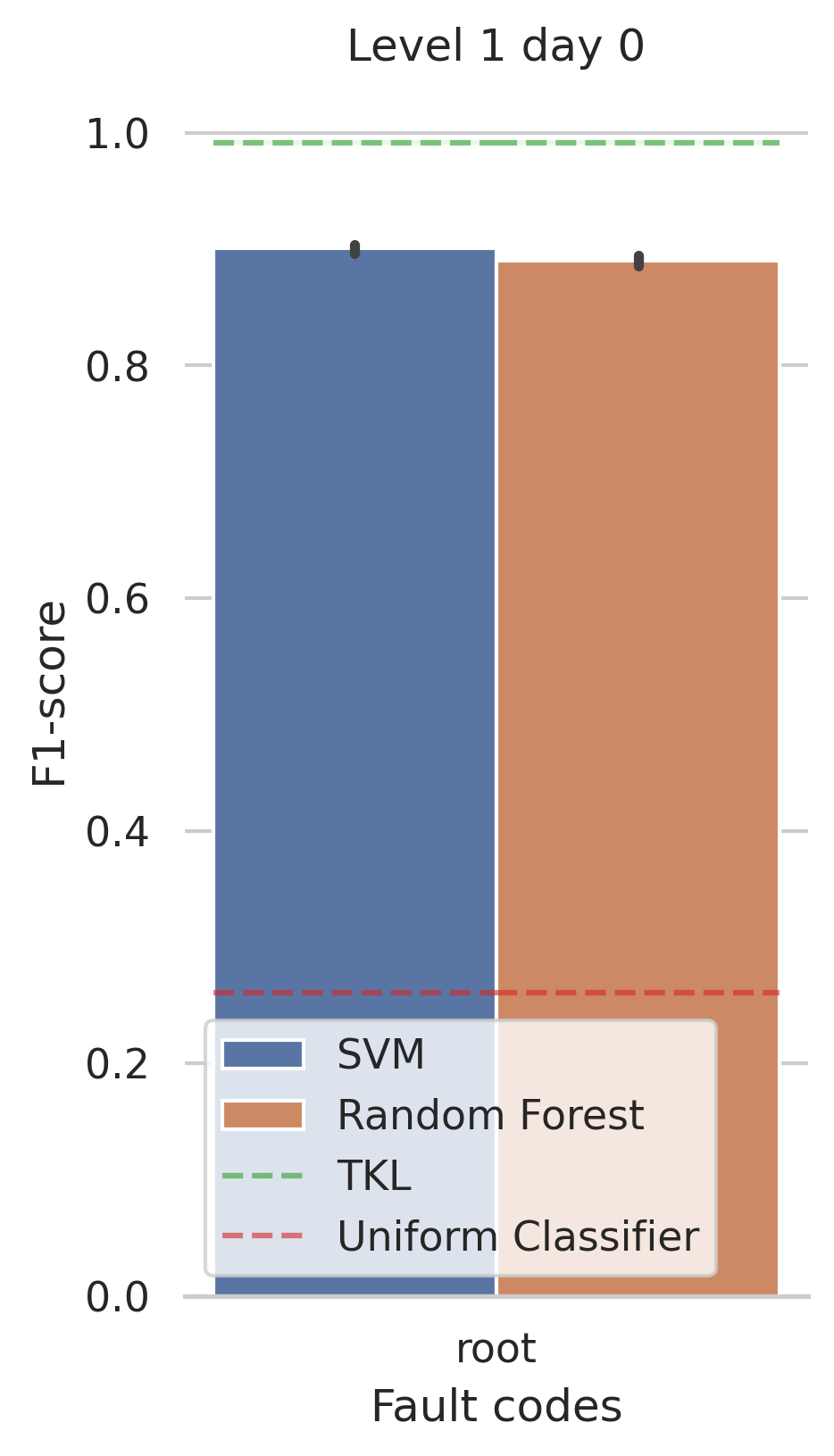}
       \includegraphics[width=0.2\textwidth,keepaspectratio]{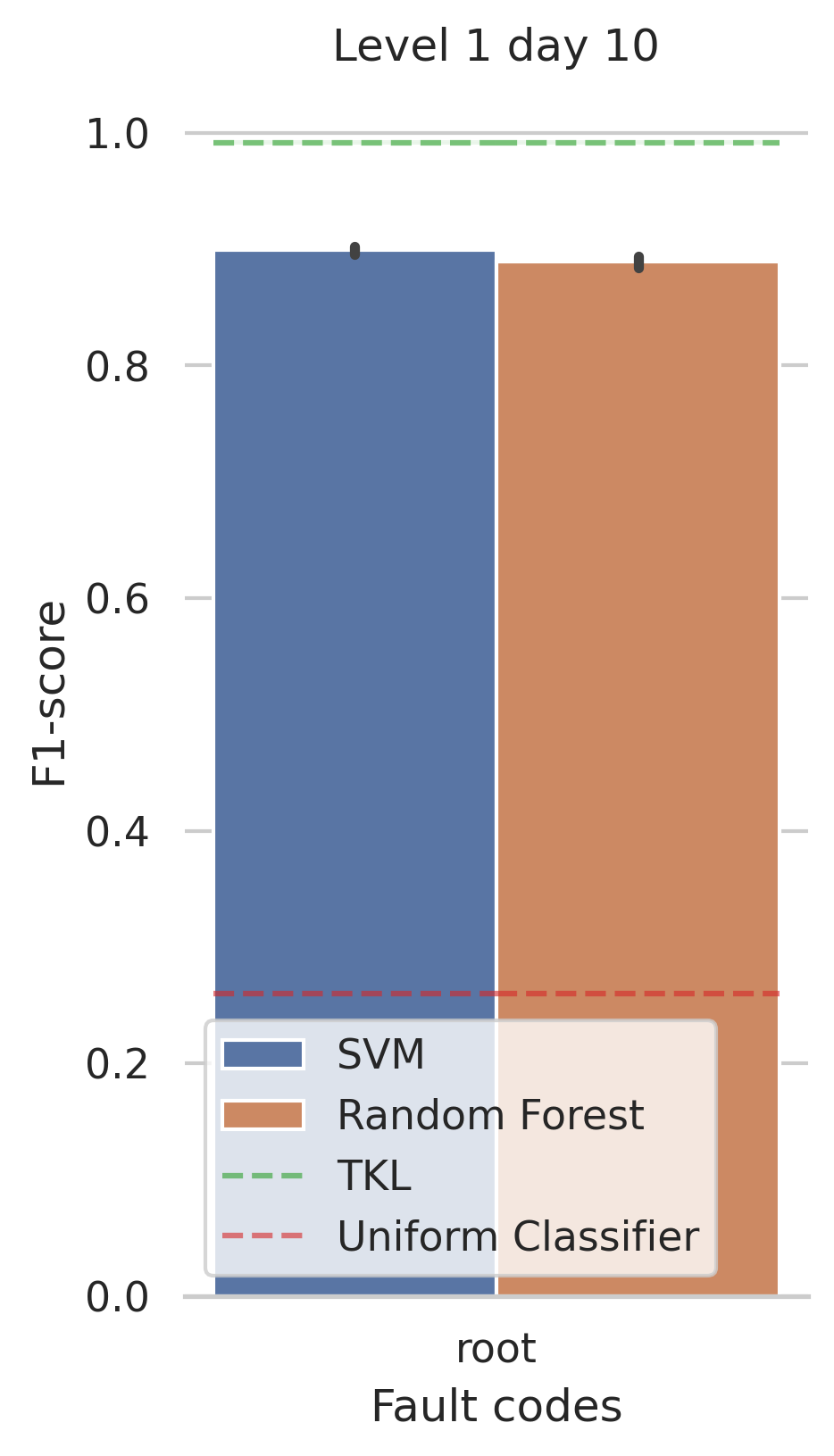}}
    \subcaptionbox{significance plot\label{fig:sp_n1_d0}}
        {\includegraphics[width=0.4\textwidth,keepaspectratio]{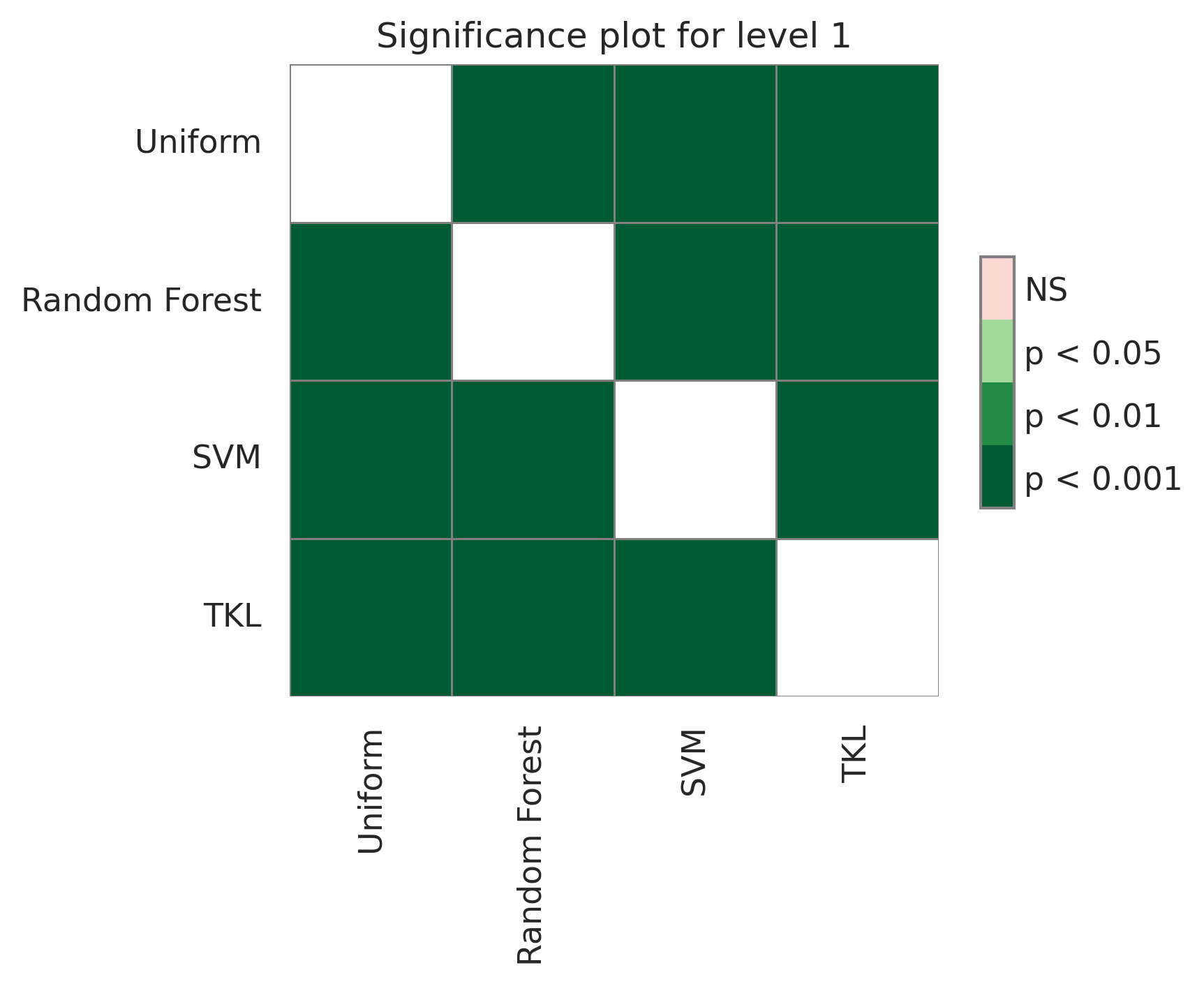}}
    \caption{Mean F1-score for the SVM and Random Forest when classifying the first level delay attribution codes, including confidence intervals. TKL and Uniform classifier are shown as lines, with confidence interval included for TKL.}
    \label{fig:f1_n1}
\end{figure}

The mean F1 score for the different algorithms are presented in Table~\ref{tab:l1}. The results indicate that the difference between Random Forest and SVM is minor, but that both algorithms perform significantly better than the uniform classifier. At Day 0, Random Forest had a mean F1-score of 0.890 (±0.008), and similarly at Day 10 with an F1-score of 0.889 (±0.008). Similarly, SVM exhibited strong performance, registering an F1-score of 0.901 (±0.006) at Day 0 and similarly at Day 10 with an F1-score of 0.899 (±0.006). The TKL classifier is the performance of the manual classification process (i.e. the F1-score calculated on the delay attribution code for day 0 compared against day 10). The results indicates that the models are capable of correctly classify the delay attribution code for level 1 using only the unstructured text from day 0 in $\approx 89\%$ of the tested instances. This is visualized in Figure~\ref{fig:f1_n1}, showing the performance against both day and day 10. This also indicates that the difference in performance between day 0 and day 10 is minimal.

A Kruskal-Wallis test showed that at there was a significant difference of means ($H = 59.338$, $p<0.001$). A Conover's post-hoc test indicates that the difference is a statistical significance between all four variables ($p<0.001$), see Table~\ref{tab:con1}. This is visualized in the significance plots in Figure~\ref{fig:sp_n1_d0}. this indicates that the median of the different samples are different from each other. Given the small standard deviation shown in Table~\ref{tab:l1}, this isn't surprising.
  
\begin{table*}[ht]
    \centering
    \footnotesize
    \begin{tabular*}{\textwidth}{@{\extracolsep\fill}lrrrr@{}}
        \toprule
        Algorithm &   Uniform Classifier & Random Forest &    SVM &    TKL \\
        Day &        &       &       &       \\
        \midrule
        0   &  $0.260$ ($0.006$)& $0.890$ ($0.008$) & $0.901$ ($0.006$) & $0.992$ ($0.003$) \\
        10  &  $0.260$ ($0.006$)& $0.889$ ($0.008$) & $0.899$ ($0.006$) & $0.992$ ($0.003$) \\
        \bottomrule
    \end{tabular*}
    \caption{Mean F1-score for the different algorithm on day 0 and 10 for the first level delay attribution codes. Standard deviation within parenthesis.}
    \label{tab:l1}
\end{table*}

\begin{table}[ht]
    \centering
    \footnotesize
    \begin{tabular}{lllll}
        \toprule
        {} & Uniform & Random Forest &    SVM &    TKL \\
        \midrule
        Uniform       &   $1.000$ &         $p<0.01$ &  $p<0.01$ &  $p<0.01$ \\
        Random Forest &   $p<0.01$ &         $1.000$ &  $p<0.01$ &  $p<0.01$ \\
        SVM           &   $p<0.01$ &         $p<0.01$ &  $1.000$ &  $p<0.01$ \\
        TKL           &   $p<0.01$ &         $p<0.01$ &  $p<0.01$ &  $1.000$ \\
        \bottomrule
    \end{tabular}
    \caption{Conover post-hoc test results for the first level delay attribution codes.}
    \label{tab:con1}
\end{table}

\subsection{Level 2}
\label{sec:level2}

\begin{figure}[ht]
    \centering
    \subcaptionbox{Day 0\label{fig:f1_n2_d0}}
       {\includegraphics[width=0.4\textwidth,keepaspectratio]{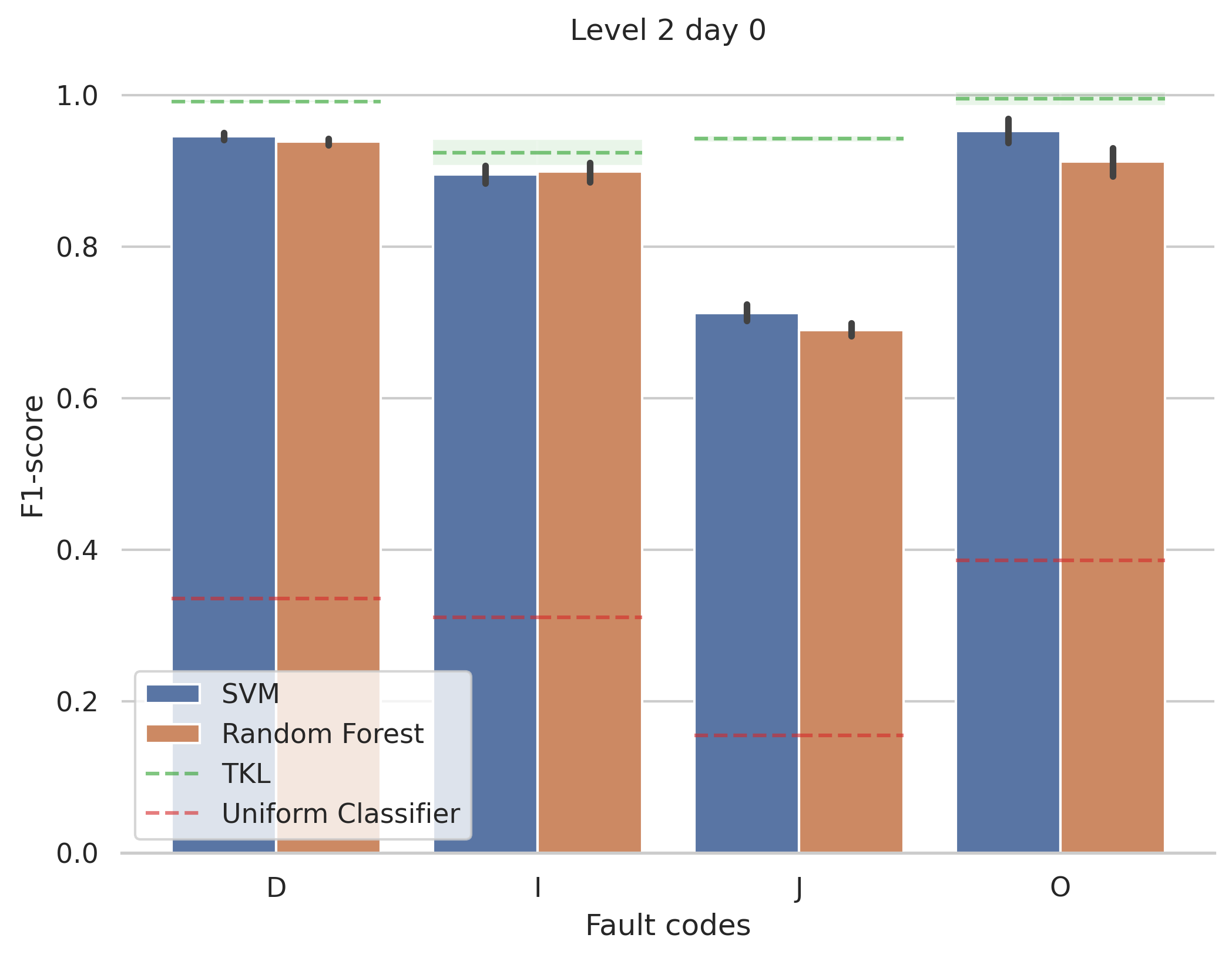}}
    \subcaptionbox{Day 10\label{fig:f1_n2_d10}}
       {\includegraphics[width=0.4\textwidth,keepaspectratio]{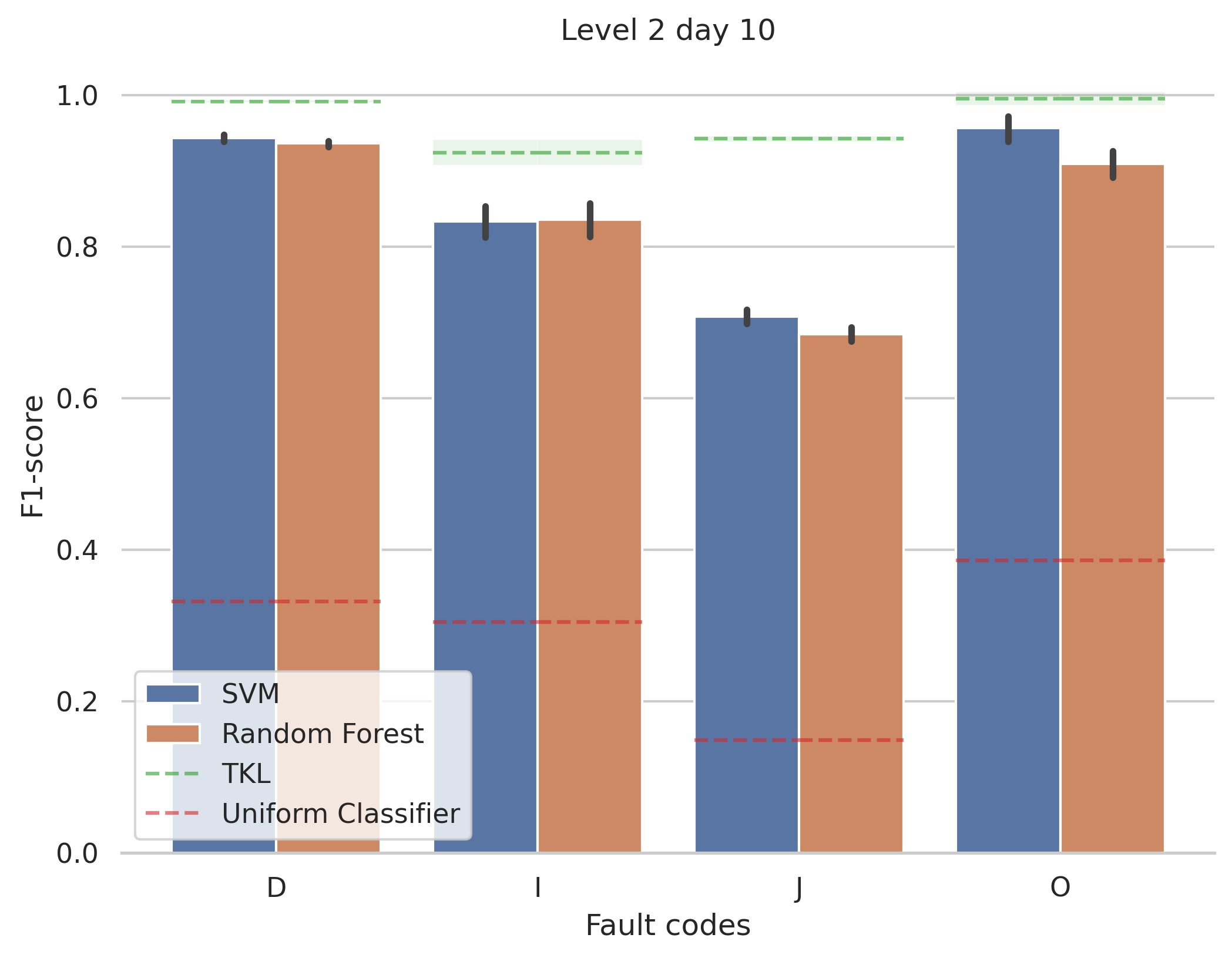}}
       
    \caption{Mean F1-score for the SVM and Random Forest when classifying the second level delay attribution codes, including confidence intervals. Classification is done per parent code, e.g. models trained and evaluated on D-codes. TKL and Uniform classifier are shown as lines, with confidence interval included for TKL.}
    \label{fig:f1_n2}
\end{figure}

The mean F1-score for the different algorithms are presented in Table~\ref{tab:l2}. The results indicate that the SVM and Random Forest performs significantly better than the Uniform Classifier, but worse than the manual classifier (TKL). In examining the performance of Random Forest and SVM across various classes (D, I, J, O) at Day 0 and Day 10, both algorithms demonstrate competitive results. Random Forest exhibits notable proficiency, achieving high F1-scores, particularly in classes like D and O, ranging from $0.938$ to $0.936$ and $0.912 -- 0.909$ respectively at the two evaluation time points. Similarly, SVM showcases commendable performance, achieving competitive F1-scores, especially in classes such as D and O, ranging from $0.946$ to $0.943$ and $0.953$ to $0.956$.  However, in the case of the delay attribution code $J$, both Random Forest and SVM performs worse than for the other delay attribution codes ($0.690$ and $0.712$ respectively at day 0, indicating that the algorithms have trouble separating some of the classes for the specific delay attribution code. The uniform classifier also have a lower mean indicating that the J code have a larger solution set (i.e. more classes) than the other delay attribution codes. This is further visualized in Figure~\ref{fig:f1_n2}

\begin{table*}[ht]
    \centering
    \footnotesize
    \begin{tabular*}{\textwidth}{@{\extracolsep\fill}llrrrr@{}}
        \toprule
        \multicolumn{2}{r}{Algorithm}  &  Uniform Classifier &    Random Forest &   SVM &   TKL \\
        Code & Day &        &       &       &       \\
        \midrule
        D & 0  &  $0.344$ ($0.007$) & $0.938$ ($0.007$) & $0.946$ ($0.007$) & $0.992$ ($0.002$) \\
          & 10 &  $0.342$ ($0.007$) & $0.936$ ($0.007$) & $0.943$ ($0.008$) & $0.992$ ($0.002$) \\
        I & 0  &  $0.359$ ($0.029$) & $0.899$ ($0.023$) & $0.895$ ($0.020$) & $0.924$ ($0.023$) \\
          & 10 &  $0.347$ ($0.028$) & $0.835$ ($0.039$) & $0.833$ ($0.036$) & $0.924$ ($0.023$) \\
        J & 0  &  $0.167$ ($0.006$) & $0.690$ ($0.015$) & $0.712$ ($0.018$) & $0.942$ ($0.005$) \\
          & 10 &  $0.168$ ($0.009$) & $0.684$ ($0.016$) & $0.707$ ($0.016$) & $0.942$ ($0.005$) \\
        O & 0  &  $0.469$ ($0.051$) & $0.912$ ($0.030$) & $0.953$ ($0.029$) & $0.995$ ($0.012$) \\
          & 10 &  $0.464$ ($0.055$) & $0.909$ ($0.030$) & $0.956$ ($0.028$) & $0.995$ ($0.012$) \\
        \bottomrule
    \end{tabular*}
    \caption{ Mean F1-score (and Standard Deviation within parenthesis) for the approaches when classifying the second level delay attribution codes.}
    \label{tab:l2}
\end{table*}

\begin{figure}[ht]
    \centering
    \includegraphics[width=\textwidth,keepaspectratio]{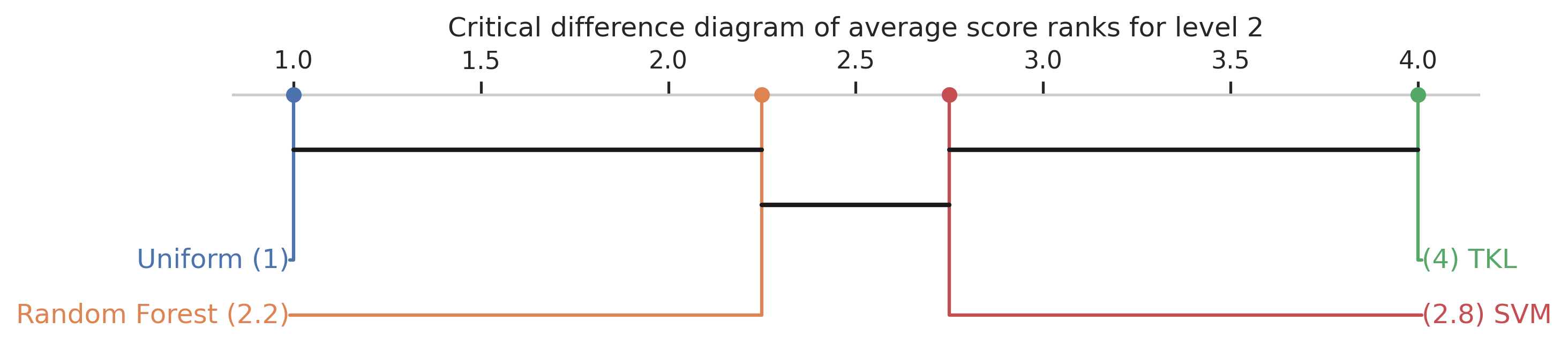}
    \caption{Critical Difference diagram for level 2 codes. The connecting line indicating no statistical difference between Random Forest and SVM, TKL and SVM, nor between Uniform Classifier and Random Forest. However, SVM is statistically significant better than the Uniform Classifier. Similarly, TKL performs better than both  the Uniform Classifier and Random Forest.}
    \label{fig:cd_n2}
\end{figure}

A Friedman test showed that a statistically significant difference was found between the algorithms, $\chi^2 (21) = 22.19, p = <0.01$. A Nemenyi post-hoc test indicates that there is a statistical significant difference between the uniform classifier and both SVM ($p<0.05$) and TKL ($p<0.01$) but not for Random Forest, see Table~\ref{tab:nem2}. Further, there is a statistical significant difference between TKL and Random Forest ($p<0.05$). The differences are visualized in the critical difference plot in Figure~\ref{fig:cd_n2}. 

\begin{table}[ht]
    \centering
    \footnotesize
    \begin{tabular}{lllll}
    \toprule
    {} & Uniform & Random Forest &    SVM &    TKL \\
    \midrule
    Uniform       &   $1.000$ &         $0.213$ &  $p<0.05$ &  $p<0.01$ \\
    Random Forest &   $0.213$ &         $1.000$ &  $0.850$ &  $p<0.05$ \\
    SVM           &   $p<0.05$ &         $0.850$ &  $1.000$ &  $0.213$ \\
    TKL           &   $p<0.01$ &         $p<0.05$ &  $0.213$ &  $1.000$ \\
    \bottomrule
    \end{tabular}
    \caption{Caption}
    \label{tab:nem2}
\end{table}

\subsection{Level 3}
\label{sec:level3}

The mean F1-score for the different algorithms for the third level of the delay attribution codes are presented in Table~\ref{tab:l3}. The results indicate that the SVM and Random Forest performs significantly better than the Uniform Classifier, but worse than the manual classifier (TKL). Random Forest and SVM demonstrate competitive performance across different delay attribution codes, with F1-scores ranging from $0.503$ to $0.898$ and $0.504$ to $0.907$, respectively. Similar to level 2, both Random Forest and SVM have difficulties when classifying most J-based delay attribution codes (e.g. JIA and JPR) compared to the other delay attribution codes, indicating that the algorithms have trouble separating some of the classes for the specific delay attribution code. The uniform classifier also have a lower mean indicating that there the J code have a larger solution set (i.e. more classes) than the other delay attribution codes. This is further visualized in Figure~\ref{fig:f1_n2}. At level 3, the overlap between different delay attribution codes can also be higher. Additionally, the dash delay attribution code ("-") is also present at level 3, acting as delay attribution code when operators cannot classify on the third level. This also increases the chances for overlap between delay attribution codes.

While most classes have similar performance between day 0 and day 10, this is not the case for the IBT code. The IBT code performs similar to the Uniform Classifier when evaluated against day 10. However, this is not unexpected as the solution set for IBT codes increases from two codes ($IBT -, IBT 40$) to six codes ($IBT -, IBT 21, IBT 22, IBT 21, IBT 30, IBT 40$). Since the model haven't been able to train against the new classes, it cannot be expected to correctly classify them either, and thus decreasing the performance. It should also be noted that the TKL performance for the IBT code is lower than the other codes, indicating that it is a difficult code to classify.

\begin{figure}[ht]
    \centering
    \subcaptionbox{Day 0\label{fig:f1_n3_d0}}
       {\includegraphics[width=0.45\textwidth,keepaspectratio]{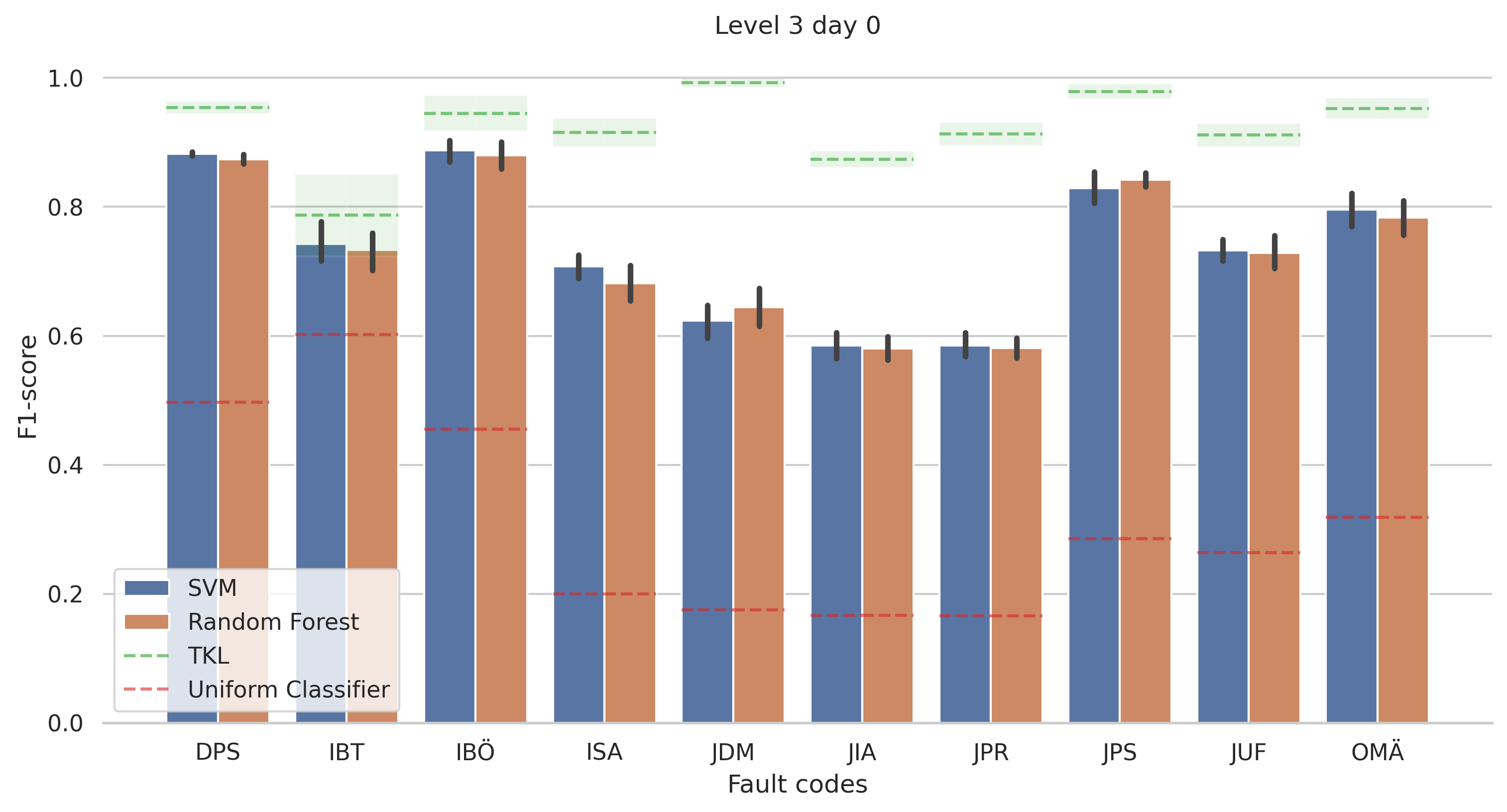}}
    \subcaptionbox{Day 10\label{fig:f1_n3_d10}}
       {\includegraphics[width=0.45\textwidth,keepaspectratio]{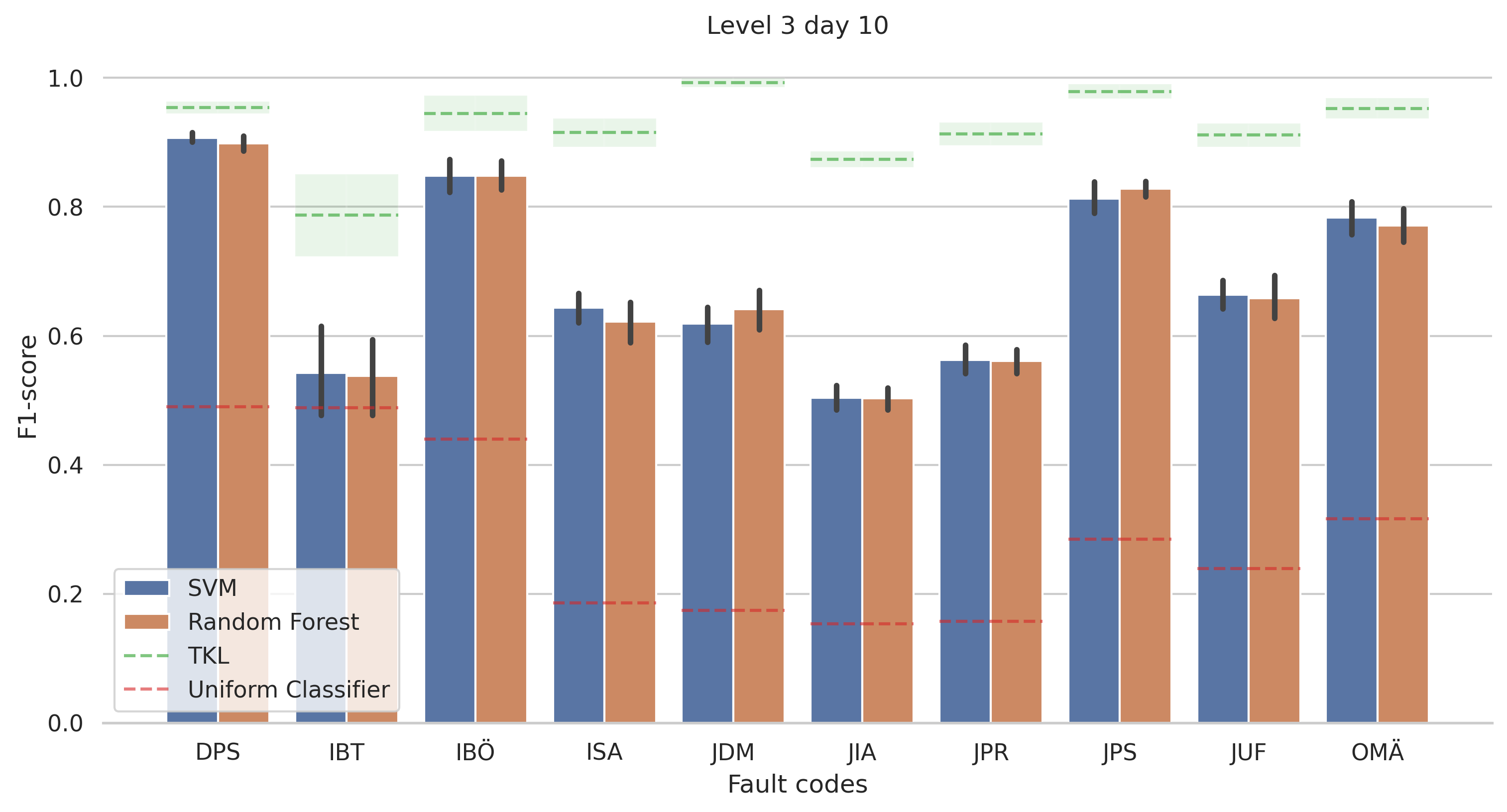}}

    \caption{Mean F1-score for the SVM and Random Forest when classifying the third level delay attribution codes, including confidence intervals. TKL and Uniform classifier are shown as lines, with confidence interval included for TKL.}
    \label{fig:f1_n3}
\end{figure}

\begin{table*}[ht]
    \centering
    \footnotesize
    \begin{tabular*}{\textwidth}{@{\extracolsep\fill}llrrrr@{}}
        \toprule
        \multicolumn{2}{r}{Algorithm}  &  Uniform Classifier &    Random Forest &   SVM &   TKL \\
        Code & Day &        &       &       &       \\
        \midrule
        DPS & 0  &  $0.497$ ($0.014$) & $0.874$ ($0.012$) & $0.882$ ($0.005$) & $0.954$ ($0.013$) \\
            & 10 &  $0.490$ ($0.015$) & $0.898$ ($0.020$) & $0.907$ ($0.013$) & $0.954$ ($0.013$) \\
        IBT & 0  &  $0.602$ ($0.106$) & $0.733$ ($0.050$) & $0.742$ ($0.051$) & $0.787$ ($0.089$) \\
            & 10 &  $0.489$ ($0.143$) & $0.538$ ($0.106$) & $0.543$ ($0.119$) & $0.787$ ($0.089$) \\
        IBÖ & 0  &  $0.455$ ($0.069$) & $0.880$ ($0.037$) & $0.888$ ($0.030$) & $0.945$ ($0.038$) \\
            & 10 &  $0.440$ ($0.060$) & $0.848$ ($0.038$) & $0.848$ ($0.044$) & $0.945$ ($0.038$) \\
        ISA & 0  &  $0.200$ ($0.032$) & $0.682$ ($0.048$) & $0.708$ ($0.031$) & $0.915$ ($0.031$) \\
            & 10 &  $0.186$ ($0.035$) & $0.622$ ($0.054$) & $0.644$ ($0.039$) & $0.915$ ($0.031$) \\
        JDM & 0  &  $0.175$ ($0.034$) & $0.645$ ($0.052$) & $0.624$ ($0.044$) & $0.992$ ($0.009$) \\
            & 10 &  $0.174$ ($0.034$) & $0.642$ ($0.053$) & $0.619$ ($0.046$) & $0.992$ ($0.009$) \\
        JIA & 0  &  $0.167$ ($0.046$) & $0.581$ ($0.032$) & $0.585$ ($0.036$) & $0.874$ ($0.017$) \\
            & 10 &  $0.153$ ($0.042$) & $0.503$ ($0.029$) & $0.504$ ($0.034$) & $0.874$ ($0.017$) \\
        JPR & 0  &  $0.165$ ($0.030$) & $0.581$ ($0.027$) & $0.585$ ($0.031$) & $0.913$ ($0.025$) \\
            & 10 &  $0.157$ ($0.030$) & $0.561$ ($0.029$) & $0.563$ ($0.036$) & $0.913$ ($0.025$) \\
        JPS & 0  &  $0.286$ ($0.043$) & $0.842$ ($0.019$) & $0.829$ ($0.041$) & $0.979$ ($0.015$) \\
            & 10 &  $0.285$ ($0.042$) & $0.828$ ($0.021$) & $0.813$ ($0.042$) & $0.979$ ($0.015$) \\
        JUF & 0  &  $0.264$ ($0.027$) & $0.729$ ($0.044$) & $0.733$ ($0.028$) & $0.911$ ($0.025$) \\
            & 10 &  $0.239$ ($0.027$) & $0.658$ ($0.054$) & $0.664$ ($0.038$) & $0.911$ ($0.025$) \\
        OMÄ & 0  &  $0.319$ ($0.081$) & $0.784$ ($0.046$) & $0.796$ ($0.045$) & $0.953$ ($0.022$) \\
            & 10 &  $0.316$ ($0.074$) & $0.771$ ($0.044$) & $0.784$ ($0.044$) & $0.953$ ($0.022$) \\
        \bottomrule
    \end{tabular*}
    \caption{ Mean F1-score (and Standard Deviation within parenthesis) for the approaches when classifying the third level delay attribution codes.}
    \label{tab:l3}
\end{table*}

\begin{figure}
    \centering
    \includegraphics[width=\textwidth,keepaspectratio]{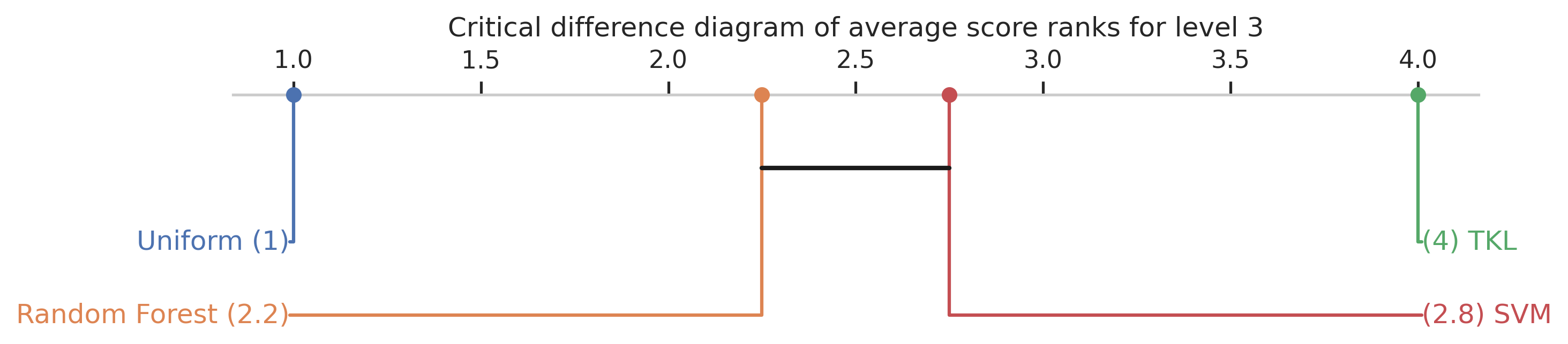}
    \caption{Critical Difference diagram for the third level delay attribution codes. The connecting line indicating not statistical difference between Random Forest and SVM, but that a statistical significant difference exists between the other approaches, e.g. TKL and SVM.}
    \label{fig:cd_n3}
\end{figure}

A Friedman test showed that a statistically significant difference was found between the algorithms over the different delay attribution codes, $\chi^2 (57) = 44.5, p = <0.01$. A Nemenyi post-hoc test indicates that there is a statistical significant difference between the uniform classifier and Random Forest ($p<0.05$), SVM ($p<0.01$) and TKL ($p<0.01$), see Table~\ref{tab:nem3}. Further, there is a statistical significant difference between TKL and SVM ($p<0.05$), Random Forest ($p<0.01$). However, no significant difference where found between SVM and Random Forest. The differences are visualized in the critical difference plot in Figure~\ref{fig:cd_n3}.

\begin{table}[h]
    \centering
    \footnotesize
    \begin{tabular}{lllll}
        \toprule
        {} & Uniform & Random Forest & SVM & TKL \\
        \midrule
        Uniform       & $1.000$ & $p<0.05$ & $p<0.01$ & $p<0.01$ \\
        Random Forest & $p<0.05$ & $1.000$ & $0.598$ & $p<0.01$ \\
        SVM           & $p<0.01$ & $0.598$ & $1.000$ & $p<0.05$ \\
        TKL           & $p<0.01$ & $p<0.01$ & $p<0.05$ & $1.000$ \\
        \bottomrule
    \end{tabular}
    \caption{Nemenyi post-hoc test for the third level delay attribution codes.}
    \label{tab:nem3}
\end{table}

\section{Discussion}
\label{sec:disc}
    In Section~\ref{sub:flatvHier}, we observe that the hierarchical models consistently outperform their flat counterparts. Specifically, the hierarchical SVM demonstrates significant improvement over the flat SVM and flat Random Forest models, while the hierarchical Random Forest outperforms the flat Random Forest. This performance gain is expected, given the reduced solution space of the hierarchical models—approximately six possible classes compared to the flat models' approximately 43. However, a trade-off emerges, evident in the higher standard deviation for the hierarchical model (Table~\ref{tab:f_vs_h}). This higher variance is anticipated, reflecting the increased variability across different classes.

    Our results affirm the viability of employing a machine learning-based approach for classifying delay types using unstructured text alone. While this approach does not surpass the expertise of operatives and relies on their input for unstructured text, it effectively identifies relevant delay attribution codes. Notably, Figure~\ref{fig:f1_n3} reveals variations in classification ease across classes, with some, like IBÖ, exhibiting clearer distinctions compared to more challenging cases like JPR. It is essential to recognize that certain classes share similarities, and their unstructured texts may overlap, especially in the third level where codes contain numeric representations or a dash ('-'), indicating a default class when operatives cannot specify a code.

    The utilization of the model should not replace operators in the process, but could serve as a valuable second opinion for operators when making decisions on assigned codes. This functionality proves particularly beneficial for new employees navigating the intricacies of the delay attribution coding process or in situations where certain codes historically exhibit a higher error rate.

    Furthermore, the model stands as a useful tool for quality evaluators, acting as an indicator in cases of disagreement between the model's predictions and operator decisions. This becomes especially relevant when the model signals a potential change in the final code on day 10. Importantly, this assumes that the code in question belongs to a class where the model has demonstrated a low classification error rate.
    
    In addition to providing a second opinion, the model can complement operators in various ways. For instance, it can assist in quickly identifying and prioritizing instances with a higher likelihood of misclassification, allowing operators to focus their attention where it matters most. Certainty estimation, facilitated by conformal prediction, adds another layer of value to the collaborative interplay between the model and operators. By offering a measure of certainty or confidence in its predictions, the model empowers operators to make more informed decisions. This becomes especially crucial in situations where the model's confidence is lower, prompting operators to exercise heightened diligence and potentially seek additional information or insights before finalizing a code assignment. The integration of certainty estimation thus enhances the decision-making process.
    
    Additionally, the model can act as a continuous learning resource for operators, aiding in the ongoing refinement of their coding decisions based on the latest patterns and trends captured by the algorithm.

    Moving forward, the practical utility of the model in operational processes warrants consideration. The analysis highlights instances where codes are infrequent, such as the absence of \emph{F} codes in the dataset, and cases like IBT codes, where models struggle due to limited instances. Notably, we observe a substantial discrepancy between day 0 and day 10 for IBT codes, likely attributed to the small sample size (200 instances). The impact of sample sizes becomes particularly evident in our evaluation. Classes with limited instances, like IBT, exhibit higher sensitivity to variations in the dataset, resulting in a pronounced effect on model performance metrics. 
    
    In situations where the available instances are sparse, the models may encounter challenges in generalizing patterns effectively. Conversely, for classes with larger sample sizes, the models benefit from a more robust training set, enabling better adaptability and yielding more stable performance across different evaluation points. Understanding the influence of sample sizes on class-specific model performance is crucial for refining the training approach. For classes with scarcity in instances, strategies such as oversampling, data augmentation, or targeted collection efforts may be considered to address the imbalance and enhance model learning. Additionally, this observation emphasizes the necessity of strategic dataset curation, ensuring representative samples across all relevant classes to foster a balanced and comprehensive model training process.

    Another noteworthy aspect discovered during the investigation is the presence of several events described by only one or two numeric values. These numerical entries often correspond to train identifications, signifying e.g. instances where one train has caused a delay in subsequent trains. This phenomenon is not confined to any specific delay attribution code. However, since the text associated with these instances comprises solely numeric values, the models encounter difficulty in interpreting the underlying meaning of the numeric values. Unlike the models, operators possess the ability to discern the significance of these numbers as they can draw on knowledge beyond the confines of the unstructured text.

    This observation underscores the critical importance of ensuring separability of features between classes and enabling the models to grasp the contextual meaning of the text. For effective machine learning, features should be designed to encapsulate distinct characteristics that allow models to discriminate between various classes accurately. In the context of our investigation, the presence of only numeric values in certain instances poses a challenge as it hampers the model's capacity to discern the nuanced differences between events. To address this limitation, there is a compelling need to augment the unstructured text with additional contextual information. This can be achieved either during the text creation phase by operators, who can provide relevant insights alongside the numeric values, or through post-processing steps involving rules-based additions to enrich the dataset.

    The ability to convey nuanced meaning to the models is important for their effective understanding and classification of diverse textual inputs. Enhancing feature separability and providing context within the unstructured text not only ensures more accurate classification but also empowers the models to generalize better across different scenarios, thereby contributing to the overall robustness and reliability of the classification system.

    An assessment of the impact of excluding instances where only numeric data was available was also done. This involved retaining approximately 13,500 instances out of the original dataset, which initially comprised around 21,500 instances. Intriguingly, the results of this evaluation revealed a minimal difference when comparing the performance of the algorithms in terms of F1-score; the models displayed similar proficiency even with the removal of these specific data points. These results are available in Appendix~\ref{app:notrain}.

    This finding prompts consideration of the models' treatment of instances with only numeric data during the training process. Notably, the limited disparity in performance suggests that the models may not be effectively incorporating or discerning patterns from instances solely comprised of numeric values. While the removal of such instances does not significantly impact overall performance metrics, it raises questions about the models' ability to generalize and derive meaningful insights from these particular data points. Further investigation into the mechanisms of feature importance and model interpretability could shed light on the extent to which these instances contribute to the overall learning process. Importantly, we recognize that this exclusion encompasses thousands of instances, each representing delay scenarios that require accurate classification, such as train delays caused by other trains. The sheer volume underscores the importance of understanding how the models handle these cases and the potential implications of their exclusion. Understanding the role of such instances in the training dynamics is essential for refining the model's capacity to discern relevant information from a diverse range of inputs, thereby enhancing its adaptability and effectiveness in real-world scenarios. As previously stated, a mitigating tactic could be to to augment the unstructured text with supplementary contextual information, particularly when the text is exclusively composed of numeric values.

    The integration of machine learning models into the code assignment process provides both consequences and benefits. While there is a risk of misclassification, especially in instances where the model lacks sufficient training data or encounters novel patterns, this could be mitigated by regular model updates, continuous training, and manual handling of edge cases. We recognize the importance of maintaining a delicate balance, avoiding over-reliance on models by encouraging operators to critically assess predictions, particularly in ambiguous situations where contextual insights are paramount. This is especially important as models may struggle with understanding context beyond the provided features, missing out on domain-specific insights known to operators.

    On the positive front, the introduction of models significantly enhances efficiency by identifying instances with a higher likelihood of misclassification, enabling operators to focus efforts where human intervention is most needed. This not only contributes to consistent decision-making but also optimizes training resources, allowing models to adapt to evolving patterns autonomously.

    Moreover, models act as invaluable second opinions for operators, and aiding in the knowledge transfer process, especially for new hires. A key benefit lies in the models' ability to estimate certainty, offering operators informed insights into the reliability of predictions. Instances of lower confidence become flags for operators to exercise caution, promoting a collaborative and cautious decision-making process.

    Ultimately, the successful integration of machine learning models into the code assignment process depends on a thoughtful balance between leveraging automation for efficiency and ensuring human expertise for nuanced decision-making. 

\section{Conclusion and Future Work}
\label{sec:con}
The results suggests the feasibility of the hierarchical classification approach, based on the performance of Support Vector Machines (SVM) and Random Forest at the different levels. The SVM algorithm, exhibiting an F1-score of $0.876$ (Level 2) and $0.737$ (Level 3), and the Random Forest algorithm, with scores of $0.860$ (Level 2) and $0.733$ (Level 3), performed statistically significantly better than their flat counterparts on Level 2. These results underscore the ability of these algorithms in capturing hierarchical relationships within delay attribution codes. The hierarchical framework, extending down to Level 3, demonstrates its potential to discern granular distinctions in delay attribution codes. The success of SVM and Random Forest in navigating the hierarchical structure suggests the broader applicability and feasibility of this approach, enabling delay attribution code classification systems in train management.

The outcomes from employing a hierarchical approach with SVM, Random Forest, compared to both manual operators, and a Uniform Random Classifier, for third-level classification indicate the feasibility of this methodology. While the train operators (TKL) stands out as the top performer, both SVM and Random Forest demonstrate statistically significant improvements compared to the Uniform Classifier. For instance, in the case of the DPS delay attribution code at Day 0, SVM achieves an F1-score of $0.882$, surpassing the Uniform Classifier's $0.497$. This signifies the potential of SVM and Random Forest to improve delay attribution code classification efficiency, even if they fall short of the operator's performance. Furthermore, our findings underscore the complexity in classifying certain delay attribution code classes, exemplified by e.g. the JPR codes on level 3, which is challenging for the SVM model and the train operators. These observations contribute valuable insights to the understanding of delay attribution code classifications, guiding future efforts to refine algorithms for improved accuracy across diverse delay scenarios.

Several avenues merit exploration to enhance the robustness and interpretability of the classification approach. Incorporating alternative language representation techniques, such as more advanced word embeddings or pre-trained transformer models like BERT or GPT, could potentially improve the model's understanding of nuanced contextual information in unstructured text. However, it should be noted that the domain specific language used in the text could be difficult to interpret for general large language models. Additionally, exploring certainty estimation methods, particularly leveraging conformal prediction, offers an avenue to quantify the model's confidence in its predictions. This could provide valuable insights into scenarios where the model may be uncertain or prone to misclassification. Moreover, investigating techniques for enhancing the explainability of the model's classifications would contribute to building trust and understanding among end-users. 

\bibliographystyle{plain}
\bibliography{references}

\appendix

\section{Detailed Results without Train indicators}
\label{app:notrain}
Several instances contained text where only a numeric value where present in the dataset. To investigate the impact of these instances, the hierarchical experiments where run without them present. The results are presented in Table~\ref{tab:l1_notrain},~\ref{tab:l2_notrain},~\ref{tab:l3_notrain}.

\begin{table}[h]
    \centering
    \footnotesize
    \begin{tabular}{lrrrr}
    \toprule
    Algorithm &   Uniform & Random Forest &    SVM &    TKL \\
    Day &        &       &       &       \\
    \midrule
    0   &  $0.247$ ($0.010$) & $0.838$ ($0.008$) & $0.850$ ($0.009$) & $0.987$ ($0.004$) \\
    10  &  $0.247$ ($0.010$) & $0.837$ ($0.010$) & $0.849$ ($0.011$) & $0.987$ ($0.004$) \\
    \bottomrule
    \end{tabular}
    \caption{Mean F1-score (and Standard Deviation within parenthesis) for the approaches when classifying the first level delay attribution codes without instances where only numeric values are present.}
    \label{tab:l1_notrain}
\end{table}

\begin{table}[h]
    \centering
    \footnotesize
    \begin{tabular}{llrrrr}
    \toprule
    \multicolumn{2}{r}{Algorithm}  &  Uniform Classifier &    Random Forest &   SVM &   TKL \\
        Code & Day &        &       &       &       \\
    \midrule
    D & 0  &  $0.342$ ($0.024$) & $0.853$ ($0.017$) & $0.861$ ($0.021$) & $0.979$ ($0.008$) \\
      & 10 &  $0.338$ ($0.022$) & $0.847$ ($0.018$) & $0.856$ ($0.019$) & $0.979$ ($0.008$) \\
    I & 0  &  $0.337$ ($0.041$) & $0.897$ ($0.023$) & $0.897$ ($0.023$) & $0.924$ ($0.026$) \\
      & 10 &  $0.319$ ($0.047$) & $0.835$ ($0.038$) & $0.835$ ($0.036$) & $0.924$ ($0.026$) \\
    J & 0  &  $0.159$ ($0.017$) & $0.682$ ($0.013$) & $0.701$ ($0.024$) & $0.943$ ($0.012$) \\
      & 10 &  $0.158$ ($0.018$) & $0.678$ ($0.013$) & $0.696$ ($0.024$) & $0.943$ ($0.012$) \\
    O & 0  &  $0.469$ ($0.051$) & $0.916$ ($0.034$) & $0.967$ ($0.023$) & $0.995$ ($0.012$) \\
      & 10 &  $0.464$ ($0.055$) & $0.912$ ($0.034$) & $0.970$ ($0.020$) & $0.995$ ($0.012$) \\
    \bottomrule
    \end{tabular}
    \caption{Mean F1-score (and Standard Deviation within parenthesis) for the approaches when classifying the second level delay attribution codes without instances where only numeric values are present.}
    \label{tab:l2_notrain}
\end{table}

\begin{table}[h]
    \centering
    \footnotesize
    \begin{tabular}{llrrrr}
        \toprule
        \multicolumn{2}{r}{Algorithm}  &  Uniform Classifier &    Random Forest &   SVM &   TKL \\
        Code & Day &        &       &       &       \\
        \midrule
        DPS & 0  &  $0.496$ ($0.020$) & $0.877$ ($0.011$) & $0.883$ ($0.006$)& $0.954$ ($0.013$) \\
            & 10 &  $0.493$ ($0.014$) & $0.897$ ($0.013$) & $0.907$ ($0.010$)& $0.954$ ($0.013$) \\
        IBT & 0  &  $0.627$ ($0.099$) & $0.749$ ($0.071$) & $0.748$ ($0.058$)& $0.803$ ($0.100$) \\
            & 10 &  $0.535$ ($0.111$) & $0.573$ ($0.118$) & $0.573$ ($0.110$)& $0.803$ ($0.100$) \\
        IBÖ & 0  &  $0.455$ ($0.069$) & $0.874$ ($0.029$) & $0.903$ ($0.034$)& $0.945$ ($0.038$) \\
            & 10 &  $0.440$ ($0.060$) & $0.843$ ($0.027$) & $0.861$ ($0.043$)& $0.945$ ($0.038$) \\
        ISA & 0  &  $0.200$ ($0.032$) & $0.685$ ($0.044$) & $0.713$ ($0.037$)& $0.915$ ($0.031$) \\
            & 10 &  $0.186$ ($0.035$) & $0.624$ ($0.054$) & $0.648$ ($0.045$)& $0.915$ ($0.031$) \\
        JDM & 0  &  $0.175$ ($0.034$) & $0.651$ ($0.046$) & $0.623$ ($0.042$)& $0.992$ ($0.009$) \\
            & 10 &  $0.174$ ($0.034$) & $0.647$ ($0.047$) & $0.618$ ($0.042$)& $0.992$ ($0.009$) \\
        JIA & 0  &  $0.152$ ($0.016$) & $0.555$ ($0.035$) & $0.571$ ($0.029$)& $0.879$ ($0.022$) \\
            & 10 &  $0.143$ ($0.019$) & $0.492$ ($0.029$) & $0.505$ ($0.030$)& $0.879$ ($0.022$) \\
        JPR & 0  &  $0.187$ ($0.030$) & $0.582$ ($0.029$) & $0.585$ ($0.027$)& $0.909$ ($0.021$) \\
            & 10 &  $0.168$ ($0.029$) & $0.561$ ($0.020$) & $0.560$ ($0.027$)& $0.909$ ($0.021$) \\
        JPS & 0  &  $0.312$ ($0.040$) & $0.868$ ($0.025$) & $0.846$ ($0.027$)& $0.979$ ($0.020$) \\
            & 10 &  $0.310$ ($0.037$) & $0.852$ ($0.024$) & $0.828$ ($0.024$)& $0.979$ ($0.020$) \\
        JUF & 0  &  $0.260$ ($0.046$) & $0.741$ ($0.039$) & $0.760$ ($0.037$)& $0.910$ ($0.028$) \\
            & 10 &  $0.235$ ($0.055$) & $0.670$ ($0.043$) & $0.689$ ($0.038$)& $0.910$ ($0.028$) \\
        OMÄ & 0  &  $0.319$ ($0.081$) & $0.792$ ($0.042$) & $0.814$ ($0.050$)& $0.953$ ($0.022$) \\
            & 10 &  $0.316$ ($0.074$) & $0.777$ ($0.039$) & $0.800$ ($0.044$)& $0.953$ ($0.022$) \\
        \bottomrule 
    \end{tabular}
    \caption{Mean F1-score (and Standard Deviation within parenthesis) for the approaches when classifying the third level delay attribution codes without instances where only numeric values are present.}
    \label{tab:l3_notrain}
\end{table}

\end{document}